\documentclass[runningheads]{llncs}
\usepackage[T1]{fontenc}
\usepackage{graphicx}
\usepackage{hyperref}
\usepackage{color}

\usepackage{xcolor}
\definecolor{darkgreen}{rgb}{0.0, 0.5, 0.0}

\usepackage{subcaption}
\usepackage{adjustbox}
\usepackage{multirow}
\usepackage{algorithm}
\usepackage{algorithmic}
\usepackage{tcolorbox}
\usepackage{caption}
\usepackage{amsmath}
\usepackage{pifont}
\usepackage{booktabs}
\usepackage{colortbl}
\usepackage{amsmath}
\usepackage{amssymb}
\usepackage{listings}
\lstdefinestyle{mypython}{
    language=Python,
    basicstyle=\ttfamily\small,
    keywordstyle=\color{blue}\bfseries,
    stringstyle=\color{darkgreen},
    commentstyle=\color{gray}\itshape,
    morekeywords={BaseModel, Field, List},
    columns=flexible,
    keepspaces=true,
    breaklines=true,
    showstringspaces=false,
    frame=single,
    rulecolor=\color{black},
    stepnumber=1,
    numbersep=5pt,
    xleftmargin=0.05\columnwidth,   
    xrightmargin=0.05\columnwidth,  
}

\begin{document}
\title{CHORUS: Zero-shot Hierarchical Retrieval and Orchestration for Generating Linear Programming Code\thanks{This paper has been accepted for presentation at the 19th Learning and Intelligent Optimization Conference (LION 19).}}
\titlerunning{CHORUS}
%
\author{Tasnim Ahmed\inst{1}\orcidID{0000-0002-0799-1180} \and
Salimur Choudhury\inst{1}\orcidID{0000-0002-3187-112X}}
\authorrunning{T. Ahmed and S. Choudhury}
%
\institute{School of Computing,
    Queen's University, Ontario, Canada\\
\email{\{tasnim.ahmed, s.choudhury\}@queensu.ca}}
\maketitle              
\begin{abstract}
Linear Programming (LP) problems aim to find the optimal solution to an objective under constraints. These problems typically require domain knowledge, mathematical skills, and programming ability, presenting significant challenges for non-experts. This study explores the efficiency of Large Language Models (LLMs) in generating solver-specific LP code. We propose CHORUS, a retrieval-augmented generation (RAG) framework for synthesizing Gurobi-based LP code from natural language problem statements.
CHORUS incorporates a hierarchical tree-like chunking strategy for theoretical contents and generates additional metadata based on code examples from documentation to facilitate self-contained, semantically coherent retrieval. Two-stage retrieval approach of CHORUS followed by cross-encoder reranking further ensures contextual relevance. Finally, expertly crafted prompt and structured parser with reasoning steps improve code generation performance significantly. Experiments on the NL4Opt-Code benchmark show that CHORUS improves the performance of open-source LLMs such as Llama3.1 (8B), Llama3.3 (70B), Phi4 (14B), Deepseek-r1 (32B), and Qwen2.5-coder (32B) by a significant margin compared to baseline and conventional RAG. It also allows these open-source LLMs to outperform or match the performance of much stronger baselines---GPT3.5 and GPT4 while requiring far fewer computational resources. Ablation studies further demonstrate the importance of expert prompting, hierarchical chunking, and structured reasoning.

\end{abstract}

\section{Introduction}
Solving Linear Programming (LP) problems, an essential field of applied mathematics, has demonstrated value in areas like supply chain management, energy scheduling, marketing, resource allocation, networking, etc \cite{zhang2024solvinggeneralnaturallanguagedescriptionoptimization}. Typically, this process involves three steps---experts use domain knowledge to translate application scenarios into problem descriptions, identifying variables, objectives, constraints, and parameters. Next, programmers encode this information using modelling languages like Python, R, or C++. Finally, automated solvers execute the optimization process to generate decisions \cite{pmlr-v220-ramamonjison23a}. This approach requires domain expertise, analytical and mathematical skills, and programming knowledge.

Large Language Models (LLMs) have become powerful tools, demonstrating remarkable performance across various tasks, including code generation and reasoning tasks \cite{dubey2024llama3herdmodels}. As LLMs grow in size, they serve effectively as standalone knowledge repositories, with facts encoded in their parameters \cite{pmlr-v202-kandpal23a}, and can be further enhanced through fine-tuning on specific downstream tasks \cite{roberts-etal-2020-much}. However, even large models often lack sufficient domain-specific knowledge for specialized tasks, and their factual accuracy can diminish as the world evolves. Moreover,  updating a model's internal knowledge through fine-tuning or pretraining remains challenging, especially with vast and frequently changing text corpora, e.g., code or tool documentation \cite{pmlr-v162-mitchell22a}. To address these limitations, retrieval techniques such as Retrieval-Augmented Generation (RAG) have been introduced \cite{NEURIPS2020_6b493230}, augmenting model inputs by appending contextually relevant documents retrieved from external knowledge corpora.
However, traditional RAG pipelines have limitations. LLMs struggle to process a large number of chunked documents (e.g., top-100), even with long-context windows, not only due to efficient constraints but also a shorter top-$k$ list (e.g., 5 or 10) usually improves generation accuracy \cite{xu2024retrieval}. With a small $k$, ensuring high recall of relevant content is crucial. A retrieval model alone may struggle with local alignments across the embedding space, while a cross-encoding ranking model better selects top-$k$ contexts from top-$N$ candidates ($N>>k$) \cite{yu2024rankrag}. Furthermore, fixed-size chunking in code documentation disrupts dependencies, breaking functions, losing syntax, and separating explanations, reducing comprehensibility and leading to ineffective retrieval with incomplete functions and missing variables, increasing hallucination risks.

Efforts to simplify mathematical programming (linear, nonlinear, etc.) using LLMs aim to enhance accessibility for non-experts. Research studies have largely focused on utilizing LLMs to generate mathematical formulations and end-to-end solutions, converting problem descriptions into solver code.
Notable contributions in generating mathematical formulation from LP problem descriptions include NL4Opt \cite{pmlr-v220-ramamonjison23a} and LM4OPT \cite{ahmed2024lm4optunveilingpotentiallarge}. The NL4Opt competition explored converting natural language into structured LP formulations. The dataset samples from this competition include an input paragraph describing an LP problem, with annotations for entity extraction and semantic parsing to generate the representation of the objective and constraints. Using the NL4Opt dataset, the LM4OPT framework was proposed to enhance the performance of smaller LLMs in generating mathematical formulations. The methodology focuses on progressive fine-tuning, starting with broader domain contexts before specializing in formulation task-specific datasets. Tsouros et al. \cite{Tsouros2023HolyG2} proposed one of the first end-to-end solutions for this task, introducing an LLM-based system that generates mathematical formulations and solver codes from prompts. While demonstrating promise on the NL4Opt dataset, it lacked benchmark comparisons and relied solely on pretrained LLMs. E-OPT \cite{yang2024benchmarkingllmsoptimizationmodeling} provided a benchmark evaluating the mathematical programming code generation capabilities of LLMs across linear, nonlinear, and tabular problems.
Teshinizi et al. \cite{AhmadiTeshnizi2023OptiMUSOM} introduced OptiMUS, a a multi-agent LLM framework to both formulate and solve mixed-integer LP problems based on natural language descriptions. It iteratively engages specialized agents, generating multiple responses per problem, leading to very high computational overhead. Another work, OptLLM \cite{zhang-etal-2024-solving} integrates LLMs with external solvers to automate the modelling and solving of LP problems. The authors curated a dataset with $100$ problems from the NL4Opt development set to evaluate the performance of GPT3.5 \cite{ye2023comprehensivecapabilityanalysisgpt3}, GPT4 \cite{openai2024gpt4technicalreport}, and their fine-tuned Qwen model. In addition, OptiGuide \cite{Li2023LargeLM} combines GPT4 with plain-text user queries to generate insights into optimization outcomes which enhances decision-making in supply chain management. Researchers have explored LLM-based approaches that solve such problems without external solvers. Yang et al. \cite{Yang2023LargeLM} introduced OPRO, a framework that iteratively refines solutions using a meta-prompt. However, LLMs are not well-suited for such tasks, as their generation relies primarily on next-token prediction rather than structured mathematical reasoning. To summarize, recent advancements have demonstrated that numerous researchers are actively working on generating problem formulations or solver code from LP or similar mathematical programming problem descriptions using LLMs. However, solver codes typically rely heavily on documentation, and popular solvers frequently update their frameworks or tool versions to incorporate new features. These solvers often function as APIs that support multiple programming languages. As a result, fine-tuning a model on a specific dataset for a particular solver ties the model to that specific solver version and programming language. A major version update in the future could render the generated codes prone to errors, necessitating further annotation and re-fine-tuning, a resource-intensive process. Thus, depending on the parametric knowledge of LLMs or fine-tuning for code generation can lead to rigid, narrowly focused solutions that require significant effort to maintain over time. Moreover, while LLMs excel in traditional analytical and algorithmic code generation due to abundant examples and libraries, the scarcity of solver-specific LP code examples limits their domain-specific knowledge, a challenge for both pretrained and fine-tuned LLMs \cite{AhmadiTeshnizi2023OptiMUSOM}.

To address this limitation, our study focuses on providing an adaptive end-to-end solution to generate solver code directly from mathematical problem descriptions, without relying on any specific solver or LLM. We hypothesize that providing an enhanced context is essential to tackle this complex task. Therefore, in this paper, we propose a novel RAG-based framework, CHORUS (\underline{C}ontextual \underline{H}ierarchical \underline{O}rchestration for \underline{R}etrieval-a\underline{U}gmented code \underline{S}ynthesis), to address this challenge.
To the best of our knowledge, this is the first work to apply a retrieval mechanism to this task. In this study, we generate LP codes for the Gurobi \cite{gurobi} solver API in Python, widely used for its efficiency and robust optimization algorithms. However, our approach remains adaptable to any mathematical problem, given an available solver. Our contributions are as follows:
\begin{enumerate}
    \item We introduce CHORUS, a novel zero-shot RAG framework that enhances LLMs' retrieval abilities by leveraging hierarchical chunking and contextual metadata from LP solver (Gurobi) documentation for code generation.
    \item Proposed two-stage retrieval process with cross-encoder reranking improves code generation accuracy significantly.
    \item Expertly crafted prompt and structured parser with reasoning steps improve the reasoning capability of LLMs as well as facilitates automated inference, execution and evaluation.
    \item Experiments on the NL4Opt-Code benchmark show significant gains over baseline LLMs and traditional RAG, allowing open-source LLMs to achieve GPT4-level performance with lower computational costs.
\end{enumerate}

\section{CHORUS}
\begin{figure}
    \centering
    \includegraphics[width=\linewidth]{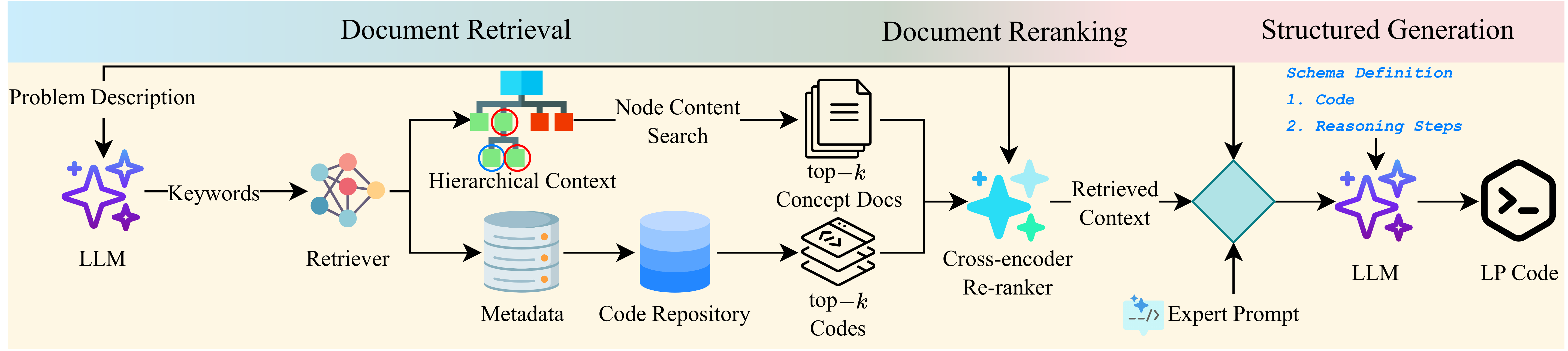}
    \caption{\textbf{CHORUS framework for generating Gurobi solver code from LP descriptions.} It retrieves and reranks relevant conceptual and implementation-level documents, supplements them with an expert prompt, and employs a structured output parser for automated execution and improved reasoning.}
    \label{fig:pipeline}
\end{figure}

This section delineates CHORUS, which enhances LLM retrieval by leveraging hierarchical chunking and contextual metadata from LP solver documentation. It describes the retrieval process with cross-encoder reranking and details the use of an expertly crafted prompt with a structured parser and reasoning steps. Figure~\ref{fig:pipeline} presents an overview of the proposed pipeline at inference.

\subsection{Document Structuring and Indexing}
The Gurobi documentation, serving as the primary source of external context, exhibits a dual nature. The first category comprises theoretical and contextual content that includes descriptions of various models, variables, and constraint types, as well as code snippets. This content is inherently rich in semantic information and provides a deep theoretical foundation necessary for understanding the nuances of different mathematical programming problem formulations. The second category consists solely of complete code examples that demonstrate the practical application of the Gurobi library in solving diverse mathematical programming scenarios. Although these code examples lack explanatory text, they embody critical implementation details that are essential for generating syntactically correct and functionally appropriate code. Given the heterogeneity in the types of content, we adopt a dual approach for storing and retrieving documents.

\subsubsection{Hierarchical Tree Indexing of Theoretical Content}
Technical documents are inherently structured, and flattening it into equal-length chunks disrupts semantic coherence. By modeling the documentation as a tree, CHORUS mimics human navigation patterns. We hypothesize that this reduces hallucination risks by ensuring the LLM receives self-contained technical context.
The theoretical documents are processed using the inherent structure of the source PDFs. We extract metadata and content from the PDF files to construct a hierarchical tree representation. At the root of this tree is the entire document, which is then segmented into chapters. Each chapter is further divided into sections, and these sections are subdivided into subsections.
The hierarchical tree is designed to preserve both detailed and high-level contextual information. The main bulk of the theoretical content is distributed across the leaf nodes, while the intermediary nodes preserve introductory texts that usually contain summaries or high-level definitions of their child nodes. Each node is considered a chunk.

\subsubsection{Metadata-Augmented Indexing of Code Examples}
Directly searching code syntax for natural language queries suffers from a vocabulary mismatch. Gao et al. \cite{gao-etal-2023-precise} introduced HyDE (Hypothetical Document Embeddings), a dense retrieval method that generates an ideal document, embeds it into a vector, and retrieves similar real documents. This approach enhances retrieval precision by capturing the conceptual essence and mitigating vocabulary mismatches. Inspired by this workflow, our proposed code example retrieval mechanism depends upon enriched metadata to bridge the semantic gap between natural language queries and stored code examples. In contrast to the theoretical contents, the complete code examples are managed as individual chunks, each representing a distinct code snippet. These snippets are stored in a vector database along with generated metadata. For every code example, we generate a set of domain-agnostic keywords ($5-7$) using an LLM providing an expert prompt. These keywords capture salient features such as the type of variables, the nature of the problem, and the specific constraints used in the model which abstract the functionality of the code. Additionally, a concise 2-3 line natural language synopsis is generated for each code snippet using the same approach. This metadata allows semantic alignment with user queries without requiring direct lexical matches to code syntax, which is critical for interdisciplinary problems where users describe constraints in non-technical terms.

\begin{figure}
    \centering
    \begin{subfigure}[b]{0.48\textwidth}
        \centering
        \includegraphics[width=\textwidth]{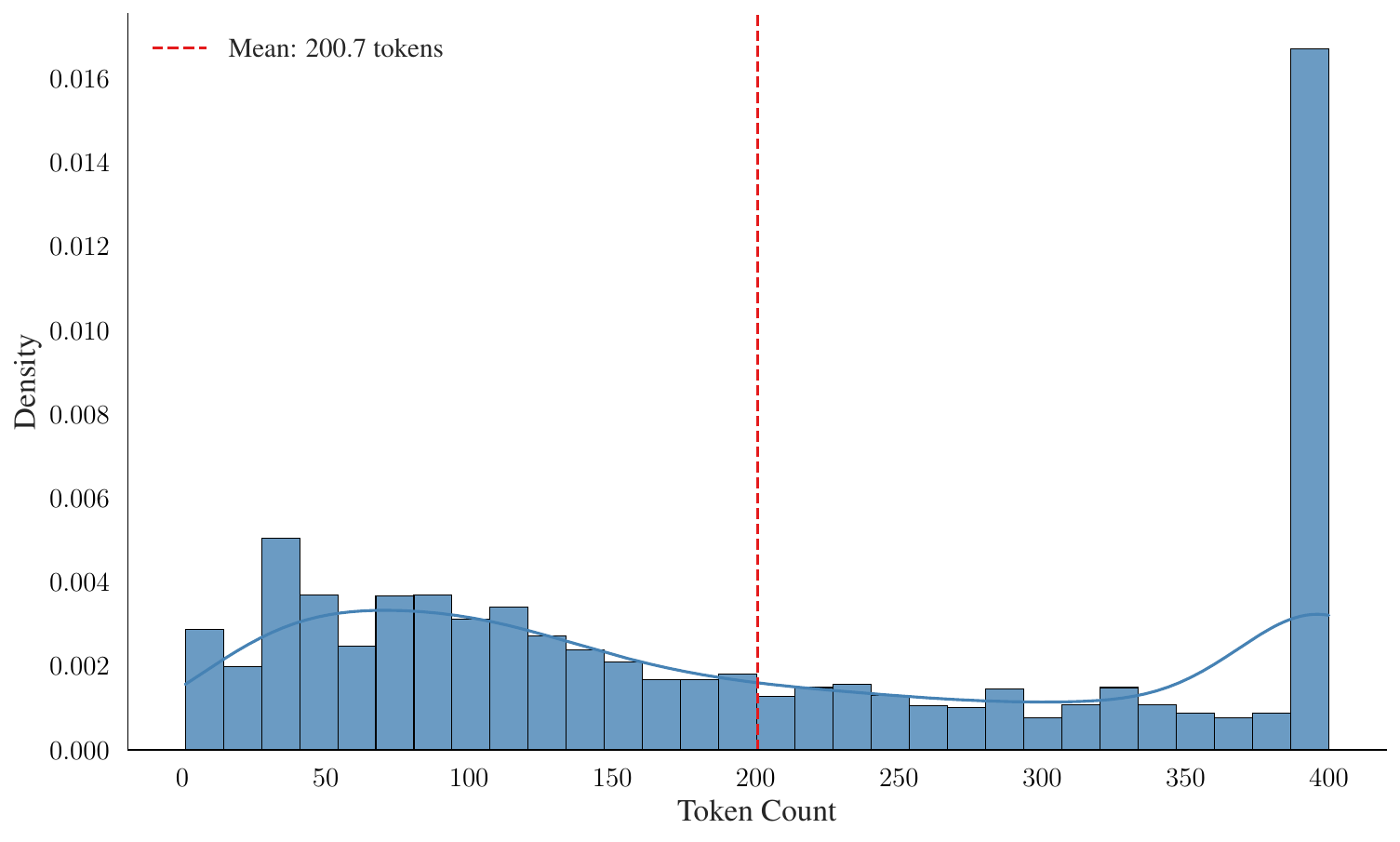}
    \end{subfigure}
    \hfill
    \begin{subfigure}[b]{0.48\textwidth}
        \centering
        \includegraphics[width=\textwidth]{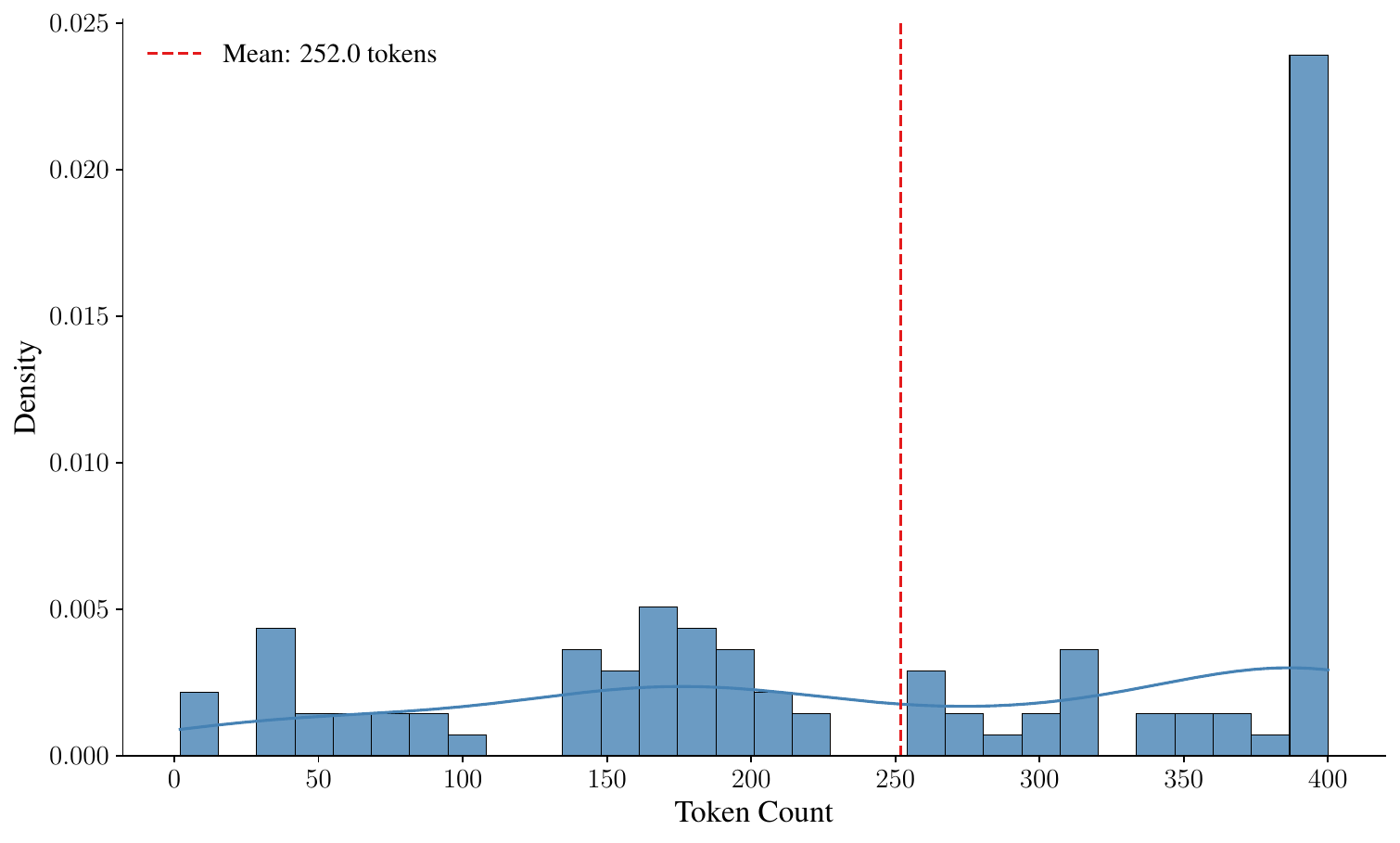}
    \end{subfigure}
    \caption{Token length distributions for (left) theoretical documentation chunks and (right) code examples.}
    \label{fig:token_distribution}
\end{figure}

\begin{figure}
    \centering
    \begin{subfigure}[b]{0.48\textwidth}
        \centering
        \includegraphics[width=\textwidth]{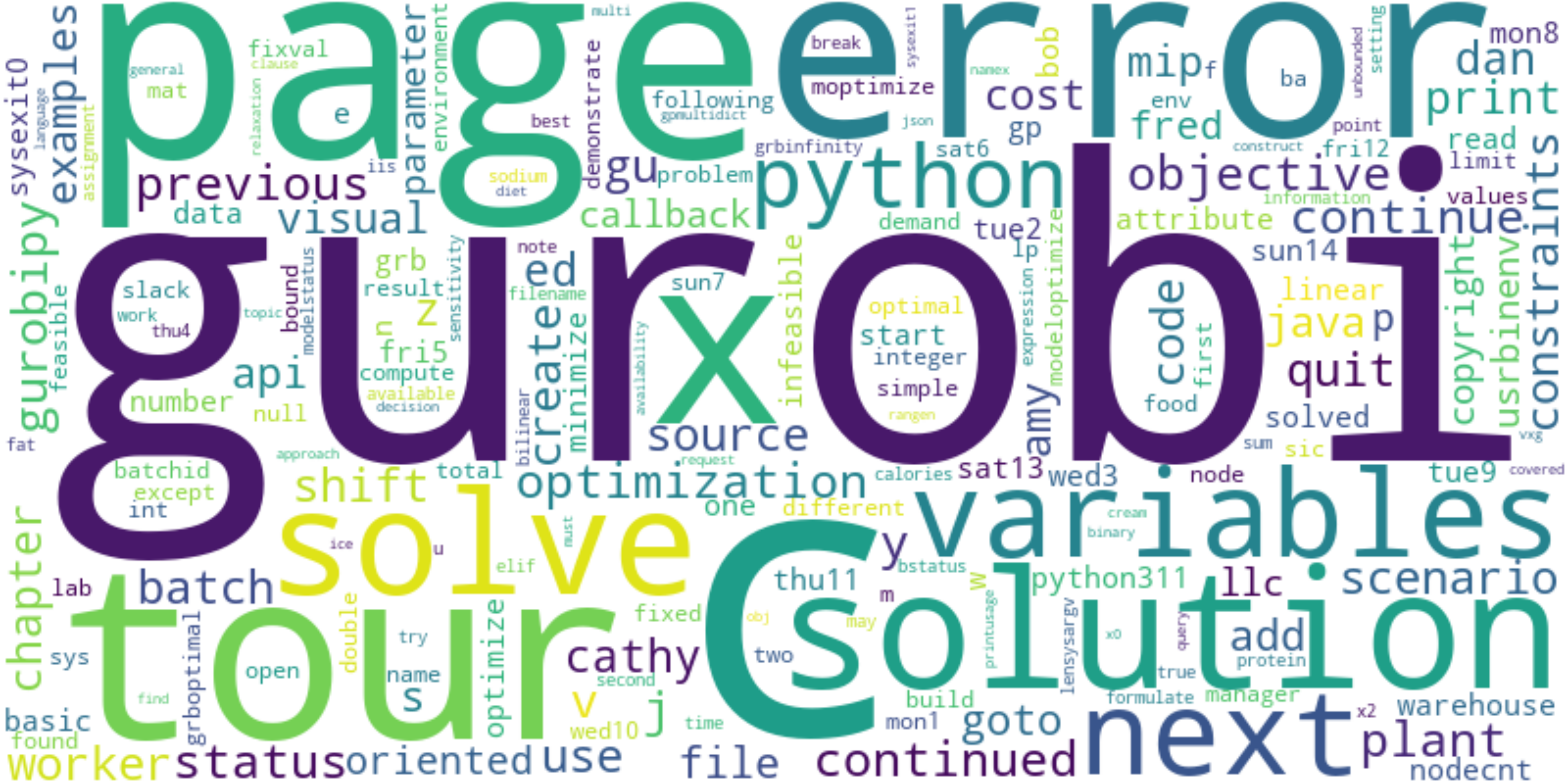}
    \end{subfigure}
    \hfill
    \begin{subfigure}[b]{0.48\textwidth}
        \centering
        \includegraphics[width=\textwidth]{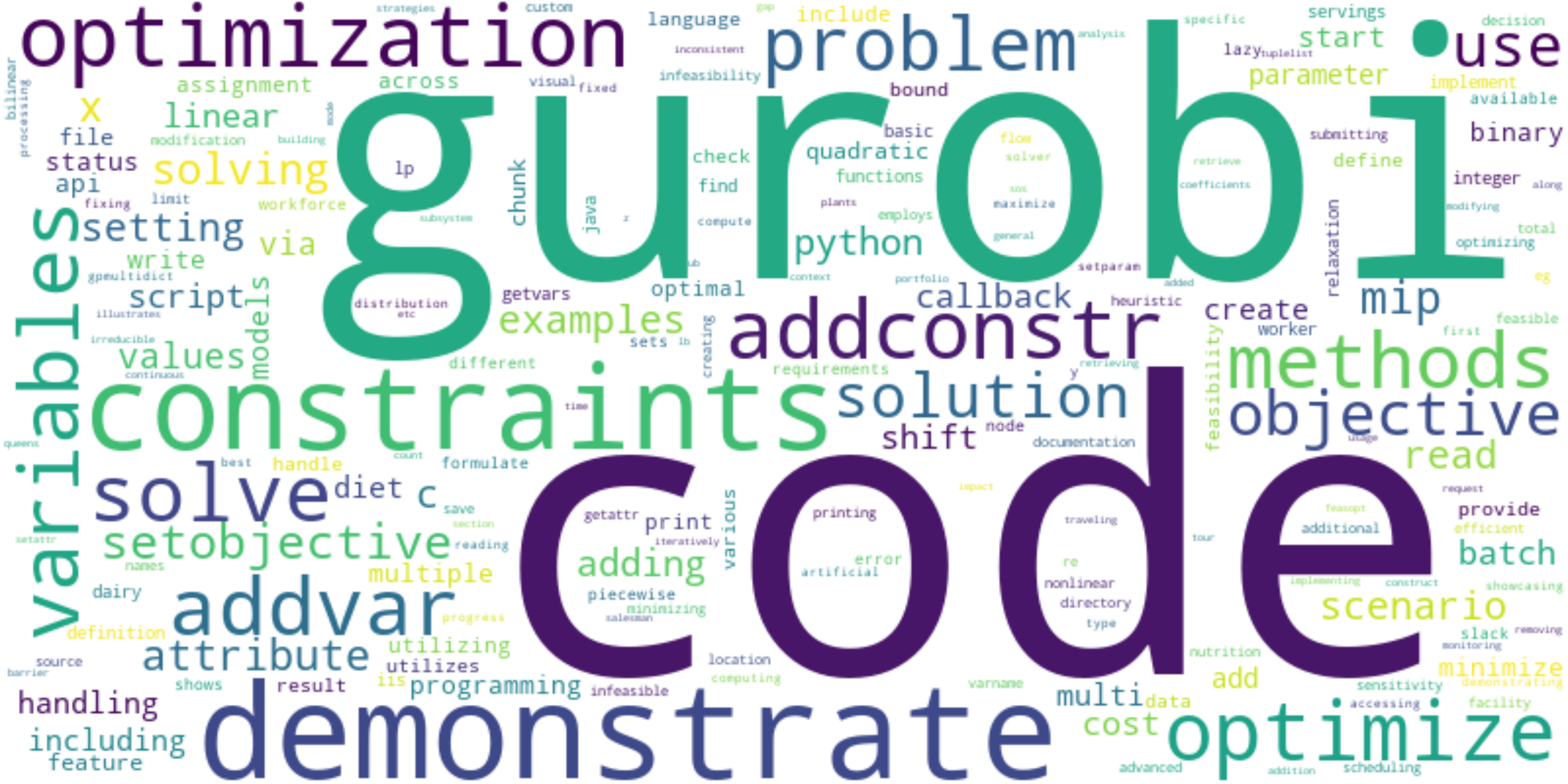}
    \end{subfigure}
    \caption{Word frequency analysis contrasting (left) raw code token prevalence with (right) curated metadata terminology from summaries and keywords.}
    \label{fig:word_clouds}
\end{figure}

The impact of our document structuring methodology is demonstrated through comparative analyses of chunk characteristics. \figureautorefname~\ref{fig:token_distribution} highlights the natural variation in chunk lengths between theoretical documents and code examples, contrasting with the rigid fixed-length chunking used in traditional RAG systems. \figureautorefname~\ref{fig:word_clouds} validates our metadata augmentation strategy through lexical analysis. The raw code term frequencies reveal dominance of implementation-specific tokens (e.g., ``gurobipy'', ``batchid'', ``continue'') that rarely appear in natural language queries. Conversely, the metadata-derived terminology demonstrates improved alignment with natural query patterns.

\subsection{Two-Stage Context Retrieval}
The retrieval process operates in two phases: conceptual context retrieval from the hierarchical documentation and code example retrieval from the indexed implementations. This separation allows the framework to retrieve both theoretical foundations and practical examples.

\subsubsection{Hierarchical Context Retrieval}
An LLM starts with extracting a set of keywords from the problem description, focusing on LP-specific concepts (e.g., ``binary variables,'' ``supply chain optimization''). Keyword extraction is guided by a prompt engineered to prioritize terms likely to appear in technical documentation. These keywords are used to traverse the hierarchical document tree. For each keyword, top-$k$ most relevant nodes are selected based on the cosine similarity between their embeddings and node content.
The retrieved chunk post-processing follows a dynamic, context-aware strategy. To ensure retrieved chunks are contextually self-contained yet concise, adaptive chunk construction dynamically combines sibling or parent content when node sizes fall below the maximum chunk size threshold ($400$ tokens). For smaller nodes, sibling content is concatenated iteratively until the token limit is reached. Crucially, parent node summaries are prepended to child content to retain hierarchical context. For example, a subsection describing quadratic constraints inherits the introductory context from its parent subsection on constraints, which ensures retrieved chunks provide both specificity and conceptual grounding. This approach mirrors human navigation patterns in technical documents, where practitioners iteratively narrow from broad overviews to granular details.

\subsubsection{Code Example Retrieval}
The second phase of the retrieval process focuses on the complete code examples. Here, the keywords extracted from the problem description are matched not directly against the code itself but against the metadata that accompanies each complete code example.
A similarity search is conducted within the vector database to retrieve top $m$ candidate examples that best match descriptive keywords. The reliance on metadata rather than raw code ensures that the retrieval process remains robust to variations in coding style and syntax. This approach avoids lexical mismatches between natural language queries (e.g., ``minimize delivery costs'') and code variables (e.g., \texttt{model.setObjective(sum(c[i,j]*x[i,j] for i,j in routes), GRB.MINIMIZE)}).

\subsection{Cross-Encoder Reranking for Contextual Relevance}
After the initial retrieval phase, the system retains two candidate sets: one comprising $k \times n_{keywords}$ conceptual documents from the hierarchical tree and the other comprising $m$ complete code examples. However, to further refine the selection and ensure that only the most relevant documents are utilized during code generation, we introduce a re-ranking mechanism based on a cross-encoder model \cite{déjean2024thoroughcomparisoncrossencodersllms}.
While dual-encoders excel at scalable retrieval, their separate encoding of queries and documents limits their ability to capture fine-grained semantic relationships.
The cross-encoder model is trained on the MS MARCO passage ranking task\footnote{\url{github.com/microsoft/MSMARCO-Passage-Ranking}}. The re-ranking process involves constructing input pairs for the cross-encoder, where each pair consists of the user query, keywords, and the text content (with metadata, in the case of code examples) of a candidate document. The cross-encoder model then computes a relevance score for each pair, effectively quantifying the semantic alignment between the query and the document. The candidate documents are subsequently sorted in descending order of their scores, and a selective truncation is performed wherein the top-$3$ conceptual documents and the top-$2$ code examples are retained for further processing.

\subsection{LLM Code Generation}
The final set of retrieved and re-ranked documents, comprising both conceptual context and code examples, is then integrated into the expert prompt that is fed to the language model. This integration serves a dual purpose---it enriches the input with domain-specific knowledge and ensures that the generated code adheres to both theoretical principles and practical implementation guidelines as prescribed by the Gurobi documentation.
\subsubsection{Expert Prompting}
To ensure that the generated code rigorously matches each problem specification while maintaining high-quality coding practices, we implement specialized system and user prompts. The system prompt directs the language model to assume the role of an expert in both operations research and Python-based Gurobi development, whereas the user prompt outlines strict requirements such as the function name, error handling strategy, variable naming conventions, and constraint modeling rules. These prompts merge specialized domain guidance with explicit style and accuracy rules, which enforce the LLM to create Python functions that accurately capture the mathematical formulation, limit unneeded output, and include exception handling for potential errors.
\subsubsection{Structured Output with Improved Reasoning}A fundamental requirement of CHORUS is its capacity for end-to-end automation, including code inference, evaluation, and execution without reliance on external annotations or human intervention. Preliminary experiments with LLMs revealed a notable inconsistency: despite configuring the sampling temperature to minimize stochasticity, model responses demonstrate significant variability across queries. This observation shows a contradictory strain between the unstructured nature of responses---rooted in their training on vast, heterogeneous corpora---and the structured precision demanded by code generation tasks. To enforce consistency, we implemented a structured output-parsing mechanism, wherein each LLM response is validated against a predefined schema (e.g., a class definition) to extract executable code. However, as demonstrated in \cite{tam-etal-2024-speak}, strict formatting constraints can impair LLMs’ performance on analytical and reasoning tasks, with degradation intensifying as constraints tighten. We assume that enforcing code-only outputs, devoid of any explanatory text, may inadvertently restrict the model’s capacity to engage in stepwise reasoning---a critical factor in ensuring code correctness. This hypothesis also aligns with empirical findings from OpenAI\footnote{\href{https://openai.com/index/introducing-structured-outputs-in-the-api/}{openai.com/index/introducing-structured-outputs-in-the-api}}, which suggest that LLM output quality improves when models are instructed to provide intermediate reasoning steps within a structured schema. Building on this insight, we augmented our output schema with a dedicated \texttt{reasoning\_steps} field, explicitly instructing the model to validate the code alignment with the problem specification (e.g., variable definitions, constraint formulations, and objective functions). The schema definition is as follows:

\begin{lstlisting}[style=mypython]
class GurobiSolution(BaseModel):
    code: str = Field(description="Complete Python code using Gurobi API that solves the LP problem")
    reasoning_steps: str = Field(description="Justification mapping code components to problem requirements")
\end{lstlisting}

\section{Experiments}
\subsection{Task Formulation}
Given a natural language problem description \( X \), the CHORUS framework implements a two-stage retrieval process to identify relevant context from structured documentation and code examples. The hierarchical retriever \( R_H \) extracts conceptual documents \( D_C = \{d_{c1}, ..., d_{ck}\} \) by traversing a tree-structured index of theoretical content, preserving parent-child context relationships. Simultaneously, the metadata-augmented retriever \( R_M \) identifies code examples \( D_E = \{d_{e1}, ..., d_{em}\} \) through keyword and summary alignment with \( X \). These candidates are reranked by a cross-encoder \( C \) to yield optimal context subsets \( D_C^* \subseteq D_C \) and \( D_E^* \subseteq D_E \). The generator \( G \), conditioned on \( X \), \( D_C^* \), and \( D_E^* \), produces structured output \( Y_{\text{structured}} \) containing:  
\[
Y_{\text{structured}} = \langle \text{code}, \text{reasoning\_steps} \rangle  
\]  
where:
\begin{itemize}
    \item \(\text{code}\): Executable Gurobi code adhering to API constraints
    \item \(\text{reasoning\_steps}\): Natural language justification mapping code to \( X \)
\end{itemize}

\subsection{Dataset and Metrics}
To the best of our knowledge, there are currently no publicly available datasets on general LP problems except the data from the NL4Opt competition. We conducted an empirical study using the NL4Opt-Code dataset, which we curated from the original problem descriptions found in the NL4Opt dataset.
The NL4Opt dataset includes LP Word Problems (LPWPs) across six domains: sales, advertising, investment, production, transportation, and sciences. Each LPWP has problem descriptions and annotations for key entities such as constraints, objectives, and variables. We expanded this dataset by generating Gurobi code for each LPWP. The NL4Opt-Code dataset now contains LPWPs, each paired with its corresponding Gurobi solver code.

\subsubsection{Dataset Annotation} The dataset annotation was performed by a domain expert---a graduate student who successfully completed an undergraduate course in Optimization and a graduate course in Combinatorial Optimization. Annotation process was carried out in two stages. In the first stage, the annotator derived the LP formulation for each problem, ensuring that the mathematical representation accurately captured the nuances of the problem description. These formulations were then verified by another expert with similar qualifications to ensure their correctness. In the second stage, based on the verified formulations, the corresponding Gurobi code was generated for the Python API.

\subsubsection{Evaluation Metrics} The evaluation metric used in E-OPT \cite{yang2024benchmarkingllmsoptimizationmodeling} uses the output code to produce numerical answers for variables and objectives, which are compared against the ground truth for accuracy. On the other hand, NL4Opt compares the coefficients of variables and decision variables in constraints and the objective function against the ground truth based on an intermediate matrix representation. However, this approach imposes certain limitations, such as requiring all constraints to be in the less-than-or-equal-to form and the objective function to be in the maximization form. As a result, a correct formulation may be considered incorrect if it does not conform to these specific rules. Unlike NL4Opt and ReSocratic, we adopted a more stringent definition for our primary evaluation metric, ``accuracy.'' In addition, we incorporated three supplementary measures that evaluate code quality at varying levels of detail—specifically, ``syntactic validity,'' ``semantic similarity,'' and ``edit distance.''

\noindent\underline{Accuracy:} A generated response is considered correct only if the optimal objective function value from the generated solver code matches that of the reference solver code. This metric offers an automated and fast method for evaluating the solution, eliminating the need for human evaluators. Additionally, when the objective value matches, it generally indicates that all constraints and objective functions are correctly aligned.

\begin{equation}
\text{Accuracy} = \frac{1}{|X|} \sum_{x \in X} \mathbb{I}\left(f( Y_{gen}(x) ) = f( Y_{ref}(x) )\right)
\end{equation}

\noindent Here,\\
- \( X \) be the set of all problem instances in the test set.\\
- \( Y_{\text{gen}}(x) \) be the generated code for problem instance \( x \in X \).\\
- \( Y_{\text{ref}}(x) \) be the reference code for problem instance \( x \in X \).\\
- \( f \) be the optimal objective function value produced by executing a solver code.\\
- \( \mathbb{I} \) be the indicator function, where \( \mathbb{I}(A) = 1 \) if condition \( A \) is true, and \( \mathbb{I}(A) = 0 \) otherwise.

\noindent\underline{Syntactic Validity:} It measures whether the generated code follows proper Python syntax rules.
\begin{equation}
\text{Syntactic Validity} = \frac{1}{|X|} \sum_{x \in X} \mathbb{I}\left(\text{parse}(Y_{gen}(x)) = \text{True}\right)
\end{equation}

\noindent\underline{Semantic Similarity:} It quantifies the conceptual similarity between the generated and reference code using embeddings from a pre-trained language model (\texttt{all-MiniLM-L6-v2\footnote{\url{huggingface.co/sentence-transformers/all-MiniLM-L6-v2}}}). It is calculated using cosine similarity between the embedded representations.
\begin{equation}
\text{Semantic Similarity} = \frac{1}{|X|} \sum_{x \in X} \cos(\text{embed}(Y_{gen}(x)), \text{embed}(Y_{ref}(x)))
\end{equation}

\noindent\underline{Edit Distance:} This similarity metric measures the character-level similarity between the generated and reference code using the Levenshtein ratio.
\begin{equation}
\text{Edit Distance} = \frac{1}{|X|} \sum_{x \in X} \text{SequenceMatcher}(Y_{gen}(x), Y_{ref}(x))
\end{equation}

\subsection{Experimental Setup}
\label{ss:exp}
We designed the evaluation framework for efficient execution on a local server. To achieve this, we intentionally excluded methodologies that depend on online context and instead relied on open-source models and retrieval processes. Some entities may prefer not to use closed-source online LLMs for security and privacy. In certain scenarios, a solution may need to run on local devices or in remote areas. For instance, a recent country-wide internet outage in a country lasted several days, and if a business or operation depends on LP solving, a standalone local solution can be crucial.

\subsubsection{Choice of LLMs}
\label{sss:llm}
We chose a diverse set of LLMs---Llama3.1 (8B), Llama3.3 (70B) \cite{dubey2024llama3herdmodels}, Phi4 (14B) \cite{abdin2024phi4technicalreport}, Deepseek-r1 (32B) \cite{deepseekai2025deepseekr1incentivizingreasoningcapability}, and Qwen2.5-coder (32B) \cite{hui2024qwen25codertechnicalreport}—and compared them with the well-known baselines, GPT3.5 and GPT4. This selection covers a wide range of parameter sizes and architectures, letting us explore the trade-offs between model complexity, computational efficiency, and code generation quality. For instance, Llama3.1 (8B) gives us insight into how lightweight models perform in resource-limited settings, while Llama3.3 (70B) serves as a benchmark for high-capacity models with stronger reasoning capabilities. Phi4 (14B) and Deepseek-r1 (32B) were selected for their unique training approaches that may enhance reasoning, and Qwen2.5-coder (32B), with its coding focus, is particularly suited for our code generation task.

\subsubsection{Hardware Configuration}
The experiments were conducted on a Linux server equipped with an Intel Xeon Platinum 8358 processor, featuring 38 MB of cache, 32 cores, and a maximum clock speed of 3.2 GHz. An NVIDIA H100 NVL GPU further accelerated the computational tasks with 95.8 GB of GPU memory. The system was supported by 251 GB of RAM.

\subsection{Results}
Table~\ref{tab:results} presents a comparative analysis of the performance of the baseline models and the CHORUS framework across multiple LLMs. While GPT3.5 and GPT4 serve as baseline references, their strong performance (achieving accuracy scores of 0.5260 and 0.6367, respectively) emphasizes the effectiveness of closed-source models in LP code generation. Furthermore, the performance of GPT3.5 and GPT4 suggests that open-source LLMs (e.g., Qwen2.5-coder (32B)) are approaching the baseline performance of commercial, large-scale models. Among the open-source LLMs, the CHORUS framework results in substantial improvements in accuracy relative to the baseline. For example, the accuracy of Llama3.3 (70B) increases from 0.2289 to 0.5675, while Phi4 (14B) improves from 0.1938 to 0.6125. Notably, even smaller or mid-range models (e.g., Llama3.1 (8B) and Deepseek-r1 (32B)) demonstrate significant performance gains, which indicates that the incorporation of retrieval and re-ranking steps effectively enhances the alignment of generated code with the correct objective function and constraints. With the exception of the smallest model in this set (Llama3.1 (8B)), all open-source LLMs surpass GPT3.5 following the integration of the CHORUS framework, achieving performance levels comparable to GPT4. Given that GPT4 is considerably more computationally expensive and energy-intensive, the results demonstrate the effectiveness of CHORUS in optimizing the performance of locally deployed, resource-efficient LLMs. Syntactic validity improves in all CHORUS‐based runs, indicating that a more coherent technical context reduces parsing errors. Results show that for the baseline open-source LLMs, the syntactic validity score ranges from 0.69 to 0.96. However, incorporating the CHORUS framework improves the score to 1.00, except for Llama models, where the score is in the range of 0.98. This demonstrates that integrating theoretical documents and code examples enhances syntactic correctness by ensuring the generated code is almost always valid. Such an improvement is particularly significant for code generation tasks, where syntactic accuracy directly impacts downstream usability and execution reliability. Likewise, semantic similarity improves with CHORUS, which implies that the integrated code examples and conceptual documentation help preserve semantic alignment between the generated and reference solutions. However, edit distance, which measures token-level similarity, is not a particularly reliable metric for evaluating code generation tasks. Our experimental results show that edit distance remains nearly similar between the baseline and CHORUS-based runs, despite significant improvements in accuracy. This is because a given problem can be implemented in multiple valid ways, leading to high edit distance scores even for correct solutions. Overall, our evaluation indicates that accuracy is the most critical metric for this task, as it closely mirrors the real-world performance of such a framework in practical applications.
\begin{table}
    \centering
    \caption{Experimental results for both Baseline and CHORUS frameworks across various LLMs. The parameter sizes of the LLMs are detailed in \subsectionautorefname~\ref{ss:exp}.}
    \label{tab:results}
    \begin{tabular}{llcccc}
    \toprule
    \textbf{Approach} & \textbf{Model} & \textbf{\shortstack{Accuracy\\$(\uparrow)$}} & \textbf{\shortstack{Syntactic\\Validity $(\uparrow)$}} & \textbf{\shortstack{Semantic\\Similarity $(\uparrow)$}} & \textbf{\shortstack{Edit\\Distance $(\uparrow)$}} \\
    \midrule
    \multirow{7}{*}{Baseline} 
        & Llama3.1       & 0.0796 & 0.8581 & 0.8365 & 0.2008 \\
        & Llama3.3      & 0.2289 & 0.6920 & 0.9092 & 0.4185 \\
        & Phi4          & 0.1938 & 0.2872 & 0.9153 & 0.3921 \\
        & Deepseek-r1   & 0.1073 & 0.9654 & 0.9186 & 0.3276 \\
        & Qwen2.5-coder & 0.4644 & 0.8618 & 0.9280 & 0.4728 \\
        & GPT3.5            & 0.5260 & 1.0000 & 0.9159 & 0.3933 \\
        & GPT4              & 0.6367 & 0.9931 & 0.8721 & 0.3565 \\
    \hline
    \multirow{5}{*}{CHORUS}
        & Llama3.1       & 0.1349 & 0.9827 & 0.7941 & 0.2019 \\
        & Llama3.3      & 0.5675 & 0.9827 & 0.9291 & 0.4255 \\
        & Phi4          & 0.6125 & 1.0000 & 0.9379 & 0.4953 \\
        & Deepseek-r1   & 0.5848 & 1.0000 & 0.9141 & 0.3946 \\
        & Qwen2.5-coder & 0.5986 & 1.0000 & 0.9367 & 0.4747 \\
    \bottomrule
    \end{tabular}
\end{table}

\subsection{Ablation Study}
We conduct a comprehensive ablation analysis to quantify the contributions of the core components of our proposed framework---expert prompting, retrieval, and structured reasoning. \tableautorefname~\ref{tab:ablation} compares five system configurations across five LLMs, which reveals critical insights into optimization code generation dynamics. All configurations use structured output schema for automated evaluation.
\subsubsection{Impact of Expert Prompting}
The efficiency of expert prompting exhibits significant model-size dependence. While the Deepseek-r1 (32B) model achieves a 386.95\% accuracy improvement when using expert prompts (rising from 10.73\% to 52.25\%), the smaller Llama3.1 (8B) model suffers a 61.3\% reduction in accuracy (7.96\% to 3.08\%). Manual analysis of generated code samples reveals that this divergence derives from differences in instruction-following capacity and parametric knowledge. Larger models demonstrate superior adherence to the implementation and formatting rules of the expert prompt. Expert prompt occupies a significant portion of the context window in smaller models, which overwhelms their limited capacity, leading to syntax errors.
\begin{table}
    \centering
    \caption{\textbf{Ablation Results (Accuracy):} Structured reasoning provides consistent gains across LLMs. Traditional RAG underperforms CHORUS by 46.14–89.33\%.}
    \label{tab:ablation}
    \begin{tabular}{lccccc}
    \toprule
    \textbf{Configuration} & \textbf{Llama3.1} & \textbf{Llama3.3} & \textbf{Phi4} & \textbf{Deepseek-r1} & \textbf{Qwen2.5-coder} \\
    \midrule
    Baseline & 0.0796 & 0.2289 & 0.1938 & 0.1073 & 0.4644 \\
    Baseline + Expert Prompt & 0.0308 & 0.4833 & 0.5536	& 0.5225 & 0.5640 \\
    Traditional RAG & 0.0144 & 0.1578 & 0.1698 & 0.1168 & 0.3224 \\
    CHORUS (w/o reasoning) & 0.0104 & 0.5497 & 0.6020 & 0.5613 & 0.5721 \\
    CHORUS & 0.1349 & 0.5675 & 0.6125 & 0.5848 & 0.5986 \\
    \bottomrule
    \end{tabular}
\end{table}

\subsubsection{Traditional RAG Limitations}
Traditional fixed-length chunking strategy reduces accuracy by 46.14–89.33\% compared to CHORUS. This reduction primarily arises from two systemic flaws inherent to traditional retrieval approaches. First, most of retrieved chunks contain fragmented or incomplete contents. Second, semantic dispersion introduces irrelevant content---the majority of the retrieved chunks mismatch problem domains, confusing models with conflicting API examples. These issues disproportionately affect smaller architectures i.e., accuracy of Llamma3.1 plummets to 1.4\% with traditional RAG versus 13.5\% in CHORUS. Even coding-specialized models like Qwen2.5-coder (32B) exhibit 46.14\% reduced accuracy, which demonstrates that domain expertise cannot compensate for poor context selection.

\subsubsection{Impact of Structured Reasoning}
Omitting the \texttt{reasoning\_steps} field from CHORUS’s structured definition schema reduces accuracy by 1.71–92.29\% across model scales. The reasoning field forces models to explicitly map problem elements to code components, a process that improves constraint coverage. This aligns with chain-of-thought \cite{10.5555/3600270.3602070} but extends its benefits to structured code synthesis through schema enforcement.

\section{Conclusion}
In this work, we introduced \textbf{CHORUS}, a RAG framework for generating Gurobi-based LP code from natural language descriptions. Our methodology ensures semantic cohesion in theoretical documentation through hierarchical chunking and bridges vocabulary mismatches with metadata-augmented retrieval of code examples. Experimental results on the NL4Opt-Code dataset demonstrate that CHORUS consistently outperforms baseline LLMs. Ablation studies further highlight the importance of core components of our proposed framework. One of the key strengths of CHORUS is its flexibility, as it remains both LLM-agnostic and solver-independent. While our experiments focus on generating LP code for Gurobi, the framework can support various mathematical problems, as long as a corresponding solver is available. By decoupling retrieval-augmentation from solver-specific logic, the use of CHORUS allows for seamless adaptation to future solver updates and emerging LLMs, ensuring continuous improvements over standard code generation methods.

Although our framework demonstrates strong performance improvement compared to conventional RAG, certain limitations persist. Code generation remains highly sensitive to prompt engineering, and smaller models struggle to fully incorporate all contextual elements within limited context windows. Additionally, the alignment of LLMs with other optimization topics (e.g., integer linear, mixed, or non-linear problems) is left for future research. Despite these constraints, CHORUS lays a strong foundation by offering a resource-efficient alternative that allows open-source LLMs to approach GPT4-level performance. We believe that further exploration into adaptive retrieval strategies, multi-step reasoning, and agentic approach will open avenues for increasingly robust and domain-adapted code synthesis in optimization workflows.

\bibliographystyle{splncs04}
\bibliography{ref}
\end{document}